\documentclass[preprint,3p,times]{elsarticle}

\usepackage{amssymb}
\usepackage{multirow}

\usepackage{lineno}
\usepackage{float}
\usepackage{graphicx,epsfig}
\usepackage{subfigure}
\usepackage{color}

\usepackage[utf8]{inputenc}
\usepackage{amsmath}
\usepackage{setspace}

\journal{Journal Name}

\begin{document}

\begin{frontmatter}



\title{Labelled network motifs reveal stylistic subtleties in written texts }


\author{Vanessa Q. Marinho$^1$, Graeme Hirst$^2$ and Diego R. Amancio$^1$}

\address{$^1$Institute of Mathematics and Computer Science, University of São Paulo, São Carlos, Brazil\\
$^2$Department of Computer Science, University of Toronto, Toronto, Canada}

\begin{abstract}
 The vast amount of data and increase of computational capacity have allowed the analysis of texts from several perspectives, including the representation of texts as complex networks. Nodes of the network represent the words, and edges represent some relationship, usually word co-occurrence. Even
though networked representations have been applied to study some tasks, such approaches are not usually combined with traditional models relying upon statistical paradigms. Because networked models are able to grasp textual patterns, we devised a hybrid classifier, called \emph{labelled motifs}, that combines the frequency of common words with small structures found in the topology of the network, known as motifs. Our approach is illustrated in two contexts, authorship attribution and translationese identification. In the former, a set of novels written by different authors is analyzed. To identify translationese, texts from the Canadian Hansard and the European parliament were classified as to original and translated instances. Our results suggest that labelled motifs are able to represent texts and it should be further explored in other tasks, such as the analysis of text complexity, language proficiency, and machine translation.
\end{abstract}


\end{frontmatter}


\onehalfspacing

\section{Introduction}

The advent of Internet has allowed immediate access to an enormous amount of texts. The need to process and analyze these texts, in the form of emails, blog posts, tweets, and news, has fostered the development of methods in a variety of natural language processing (NLP) tasks, such as automatic summarization, authorship attribution, machine translation, sentiment analysis and others. Commonly used in some methods, word frequency is a simple, yet useful attribute employed to address some of these tasks~\cite{Koppel2004,Stamatatos}. In many contexts, however, the use of this attribute alone has not led to optimized results. Even when frequency attributes yield good performance, the robustness of classification systems might be undermined~\cite{BrennanG09}.  In the case of the authorship attribution task, for example, several works have reported excellent results when word frequency and other simple features are taken into account~\cite{grieve2007,Koppel:2009,Stamatatos}. However, recent works have shown that such features are prone to manipulation, as simple word statistics patterns can be easily mimicked by authors trying to conceal their identities~{\cite{BrennanG09}}. This drawback to the use of simple frequency counts in some NLP applications paves the way for the exploration of novel informative textual features, so as to provide both performance and robustness to the problems addressed.  In this scenario, some network approaches have been proposed to analyze texts using a topological point of view~\cite{cong2014approaching,Amancio2012scientometrics}.

In recent years, network theory has drawn the attention of a myriad of scientists from distinct research areas~\cite{Newman2010,SerraScirep,XinPiano,Ferretti2017}.
%
Of particular interest to the aims of this paper, networks have also been applied as a complementary tool in text analysis~\cite{cong2014approaching,semantical,Mihalcea}. A well known model, the co-occurrence (or word adjacency) representation has been extensively used in the study of text complexity, machine translations, stylometry, and disease diagnosis~\cite{Cancho2001,cong2014approaching,Mihalcea}. In this model, words are modelled as nodes in the network while the edges may represent syntactic~\cite{Cech20113614}, semantic~\cite{semantical}, or empirical~\cite{empirical} relationships. The complementary role played by co-occurrence networks in text analysis stems from their ability in considering both meso- and large-scale structure of texts, a feature markedly overlooked by bag-of-word models~\cite{cong2014approaching}.
The structure of a text is typically analyzed in terms of topological measurements~\cite{Newman2010,Costa2007}, with reinterpretations in the context of text analysis~\cite{Amancio2011a}.

While much study has been devoted to create text analysis techniques based either on statistical or networked representation and characterization, only a few works have probed the benefits of combining such distinct paradigms. For this reason, the main goal of this paper is to combine networked representations with the frequency of words. In order to do so, we explore the concept of \emph{network motif} to complement the information provided by frequency statistics in text analysis. In the current study, the combination of frequency and local structure as attributes for words is accomplished by considering node labels in each distinct subgraph. To illustrate the effectiveness of the proposed method in text analysis, we tackle the problems of identifying the authorship of texts, known as authorship attribution, and the identification of translationese. In the latter, the goal is to distinguish content originally produced in a language from content translated into that language.
As we shall show, our approach is able to represent texts in a more adequate and accurate manner. This is particularly clear when we compare the performance of the traditional approaches (based either on network or textual features) with the performance of the proposed technique, which is based on the combination of both textual and network features.


This paper is organized as follows: Section~\ref{related_work} briefly describes related work in the field of complex networks. Next, we explain our methodology in Section~\ref{methods}. A case study and the results of our hybrid classifier in the contexts of authorship attribution and translationese identification are presented in Section~\ref{results}. Finally, Section~\ref{conclusion} presents a summary of our paper and the perspectives for further studies.

\section{Related Work}\label{related_work}

%
Methods and concepts from complex networks have been successfully applied to analyze written texts. In several studies, texts are modeled as co-occurrence (or word adjacency) networks, where nodes represent distinct words and edges connect adjacent words. Co-occurrence networks have already been used to identify the authorship of texts~\cite{Mehri,Lahiri,Amancio2015AR,Marinho2016BRACIS,10.1371/journal.pone.0136076},
to distinguish prose from poetry~\cite{Roxas}, and to discriminate informative and imaginative documents~\cite{Arruda16}. Moreover, the structure of these networks has also proven useful to discriminate word senses~\cite{SilvaWordSense}. After modelling texts as co-occurrence networks, most of the approaches extract several network measurements in order to characterize the topology of the networks~\cite{Costa2007}. While most of these measurements are able to provide significant performance of the studied task, in most of the studies the textual context is disregarded after the network is obtained -- i.e. semantic elements are not fully considered in the analysis. Because
node labels (i.e. concepts associated with each node) may also play a complementary role in the analysis, the study of strategies for combining structure and semantics becomes relevant. While the combination of distinct sources of information in classification problems has been greatly investigated by the pattern recognition community for several years, such methods do not consider the particularities of each complex system under analysis. Here we take the view that textual structure and semantics can be easily combined via extraction of motifs.

In network theory, recurrent subgraphs (or \emph{motifs})  are used in a large number of applications~\cite{AmancioVoynich, Krumov_motifs, FiqiPA11, Marinho2016BRACIS, Milo2, Cabatbat, Biemann2012}. Usually, recurrent motifs are those whose frequency is larger than the expected (in a null model). These recurrent motifs are responsible for particular functions in biological and social networks~\cite{Milo,Milo2,Kashtan}.   In textual networks, motifs have also been employed to extract relevant information.
Milo et al.~\cite{Milo2} analyzed texts written in four different languages, namely English, French, Japanese, and Spanish. They observed that their respective word adjacency networks have similar motif sets. In a similar approach, Cabatbat et al.~\cite{Cabatbat} compared co-occurrence networks based on translations of the Bible and the Universal Declaration of Human Rights. They found that the frequency distribution of motifs is preserved across translations.  El-Fiqi et al.~\cite{FiqiPA11} used motifs to detect and identify the translator for the meanings of the Holy Qur'an. Their proposed method {identified the corresponding translators of the texts} with  an accuracy of 70\%. Biemann et al.~\cite{Biemann2012} extracted the frequency of directed and undirected motifs from texts in six languages to successfully distinguish human-generated texts from those generated with $n$-gram models. Amancio et al.~\cite{AmancioVoynich} analyzed the connectivity patterns in textual networks and found that the frequency distribution of motifs in real texts is uneven. According to their results, some motifs rarely occur in natural language texts. Marinho et al.~\cite{Marinho2016BRACIS} extracted the frequency of all directed motifs comprising three nodes to reveal the authorship of several books. In their experiments, the authorship was correctly assigned for almost 60\% of the books using only a small set of attributes.

While the characterization by network motifs has already been used in the context of text analysis, there is no systematic evaluation of the benefits in considering node labels in such structures. For this reason, our study focuses on devising strategies to combine structure and semantics in an effective way.

\section{Methodology}\label{methods}


In this section, we describe the creation of networks from raw texts. We also detail the proposed approach to characterize texts in terms of their semantics and structure.

\subsection{From texts to networks}

There are some pre-processing steps that can be performed prior to the creation of the co-occurrence networks, such as the removal of punctuation marks, the lemmatization of words, and the removal of function words. In this paper, we lower-cased the words and removed numbers and punctuation marks from the texts. Table~\ref{pre-processing-example} presents an extract before and after the pre-processing steps.
\begin{table}
\caption{Example of an extract after the pre-processing steps. The sentences are from the book \emph{Hard Times} written by Charles Dickens.}\label{pre-processing-example}
\centering
\begin{tabular}{{l}{l}}
\hline
\textbf{Original extract}  	& {NOW, what I want is, Facts.  Teach these}\\
 &boys and girls nothing but Facts.\\
\hline
 \textbf{Pre-processed extract} 		& {now what i want is facts teach these}\\
 & boys and girls nothing but facts\\
 \hline
\end{tabular}
\end{table}

\begin{figure}
\centering
\includegraphics[width=0.7\linewidth]{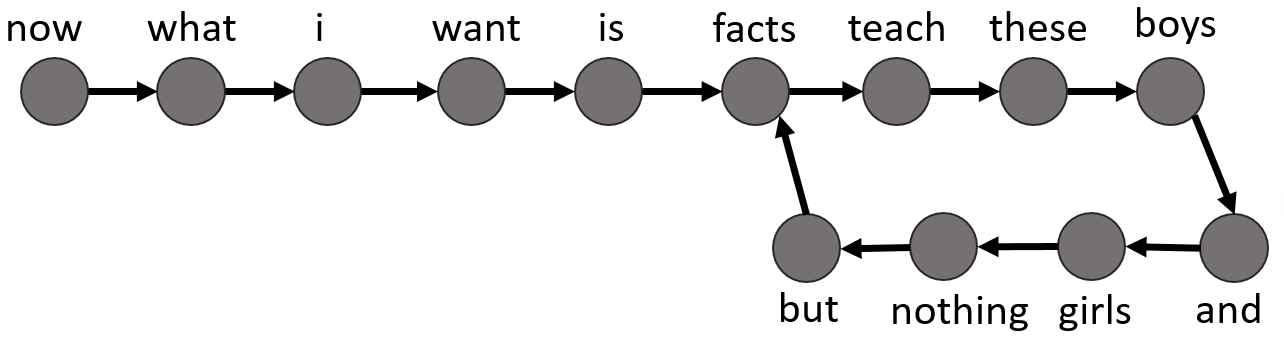}
\caption{Co-occurrence network from the pre-processed extract presented in Table~\ref{pre-processing-example}.}
\label{co-occurrence}
\end{figure}

A word co-occurrence network can be represented by a directed graph $G =
(V, A)$, where $V$ and $A$ are the set of nodes and edges, respectively. Each node $v_i \in V$ denotes a word from the vocabulary of the pre-processed text. Two vertices
are connected by an arc $a \in A$ if the corresponding words are adjacent in at least one sentence. The direction of an arc is from the first to the following word. Here, we disregarded sentence and paragraph boundaries while determining the adjacent words. Therefore, the last word of a sentence or paragraph is connected to the first word of the next sentence or paragraph. Figure~\ref{co-occurrence} presents the co-occurrence network obtained from the sentences in Table~\ref{pre-processing-example}.

\subsection{Characterization via labelled motifs}\label{characterization}

\begin{figure}
\centering
\includegraphics[width=0.85\linewidth]{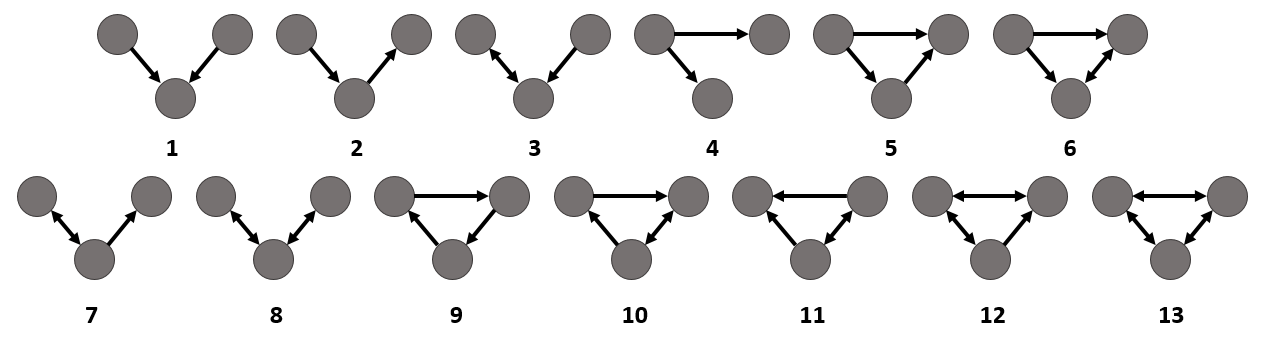}
\caption{All thirteen possible directed motifs involving three nodes.}
\label{motifs}
\end{figure}

The topology of a complex network can be characterized by several metrics. One of these metrics is the absolute
frequency of all directed motifs involving a few nodes. The set of directed motifs with three nodes is shown in Figure~\ref{motifs}. This type of representation has been used in several characterizations of complex systems~\cite{m1,m2}. Because we are interested in analyzing texts (i.e. networks with relevant information stored in each node),
we introduce the concept of \emph{labelled motifs} to take into account the information of the node labels in the subgraphs considered.

Labelled motifs are used in the strategy adopted to characterize texts by considering their frequency of appearance in the respective text networks. Instead of considering the frequency of traditional motifs, we extracted the relative frequency of a given word $w$ in each one of the 13 directed motifs displayed in Figure \ref{motifs}. More specifically, the frequency ($n_{w,m}$) of a labelled motif that combines word $w$ and motif $m$ is given by:
\begin{equation}
	n_{w,m} = \frac{\tilde{n}_{w,m}}{n_m},
\end{equation}
where $\tilde{n}_{w,m}$ is the total number of occurrences of word $w$ in motif $m$ and $n_m$ is the total number of occurrences of motif $m$, irrespective of any node labels.
Here we considered $w$ as being a word from the set of the most frequent words $W$ from the training dataset. This is the first version (V1) of the proposed method. We selected the most frequent words because {they are usually useful to characterize writing styles~\cite{Garcia_fw,AmancioVoynich}}.

\begin{figure}
\centering
\includegraphics[width=0.5\linewidth]{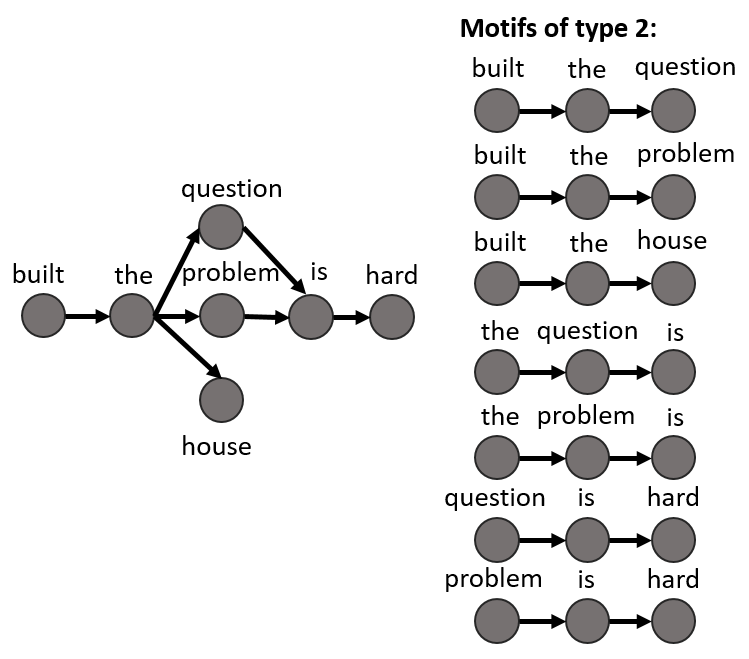}
\caption{An example of a co-occurrence network is presented on the left. On the right, we show all motifs of type 2 extracted from this network, with a total of 7, i.e. $n_m=7$. }
\label{motifs_explanation}
\end{figure}

In the version V1, a word may appear in any of the three nodes forming a motif.
To take into consideration the possibility of a word appearing in different nodes of the same network motif, we also considered the word position inside the motif according to the different configurations of nodes -- this is referred to as second version (V2).
Note that, in this version, some motif types may have equivalent nodes: this is the case of border nodes in motif type 1 and all nodes in motif type 9). In these symmetrical cases, we considered only one configuration, in order to avoid duplicated features. Examples of possible features for the toy network depicted in Figure \ref{motifs_explanation}  are described below:
\begin{itemize}
\item V1: The frequency of word \emph{the} in Motif 2. For example, if we extract labelled motifs from the network presented in Figure~\ref{motifs_explanation}, the word \emph{the} is one of the nodes in 5 occurrences of Motif type 2, i.e. $\tilde{n}_{\textrm{`the'},2}=5$. Motif 2 occurs 7 times ($n_m=7$), therefore, the frequency of word \emph{the} in Motif 2 is ${n}_{\textrm{`the'},2}=5/7$.
\item V2: The frequency of word \emph{the} as the central node in Motif 2. In Figure~\ref{motifs_explanation}, the word \emph{the} appears three times as the central node in Motif type 2, i.e. $\tilde{n}_{\textrm{`the'},2}=3$. Therefore, the frequency of word \emph{the} in this node configuration in Motif 2 is ${n}_{\textrm{`the'},2}=3/7$.
In a similar fashion, the frequency of word \emph{the} as the node with in-degree equals to 0 and out-degree equals to 1 is ${n}_{\textrm{`the'},2}=2/7$.
\end{itemize}

\subsection{Machine Learning Methods}\label{machinelearning}
In order to evaluate the performance of the proposed technique to identify stylistic subtleties in texts, we employed four machine learning algorithms to induce classifiers from a training dataset. The techniques are Decision Tree, kNN, Naive Bayes, and Support Vector Machines~\cite{Duda}. We did not change the default parameters of these methods, because comparative studies have shown that the default parameters yield, in most cases, near-optimal  results. We used a cross-validation technique with 10 folds, in which one tenth of the texts are used as a test set whereas the other nine tenths are used in the training process.

\section{Results}\label{results}
This section describes a qualitative and quantitative analysis of our method. First, in the context of the authorship attribution task, we present a case study in which \emph{labelled motifs} are used to characterize novels written by the Bront\"e sisters, known to have very similar writing styles. We also use our method to extract features in a typical authorship attribution task. In order to identify the authorship of several novels, we used two different datasets with several books from different authors. We also employed \emph{labelled motifs} in the identification of translationese. The goal of this experiment was to evaluate whether a text was originally produced in its language or it was translated into that language. In the tables presented in this section, we use the following terminology: \emph{LMV1} and \emph{LMV2} denote the results obtained with the versions V1 and V2 of our proposed technique, respectively. We also compare our results with the ones obtained with the frequency of the most frequent words. This is denoted as \emph{MFW}. The number of words used in each experiment is represented by $|W|$.

\subsection{Authorship attribution task}

Methods of authorship attribution identify the most likely author of a text whose authorship is unknown or disputed~\cite{Stamatatos}. These texts could be email messages, blog posts, or literary works, such as books and poetry.
The authorship attribution problem was the first NLP task to which we applied our method. Typically, the frequency of function words is informative as features for the problem, as noted in several works~\cite{Garcia_fw,grieve2007,Stamatatos,Koppel:2009}. However, in specific cases, these features might not perform well to distinguish very similar writing styles. This disadvantage can be overcome with our proposed technique, as we illustrate below.

In order to illustrate the ability of \emph{labelled motifs} in discriminating texts with subtle differences in style, we selected the following books: \emph{Agnes Grey} and \emph{The Tenant of Wildfell Hall}, written by Anne Bront\"e,
\emph{Jane Eyre} and \emph{The Professor} from Charlotte Bront\"e, and \emph{Wuthering Heights} from Emily Bront\"e. According to Koppel et al.~\cite{Koppel2004}, all three sisters are very hard to distinguish. In addition, by considering a dataset of books written by the three sisters, we guarantee  that all authors share the same period of time, gender, and similar education when we compare them. Therefore, the differences among the Bront\"e sisters arise from their individual writing styles~\cite{Hirst2007,Gamon2004}.

\begin{figure}
\centering
{\includegraphics[width=1\linewidth]{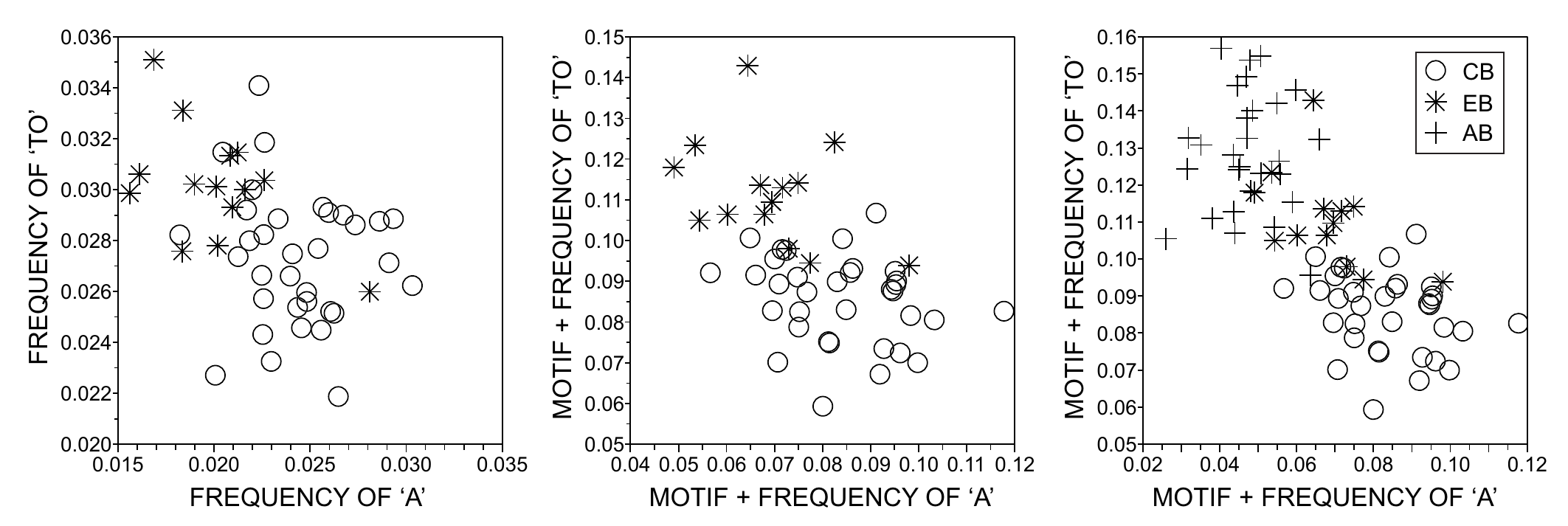}}
\caption{\label{fig:1.1}Two different feature sets are extracted from 76 partitions from books of Anne Bront\"e (AB), Charlotte Bront\"e (CB) and Emily  Bront\"e (EB). In (a), the partitions from books of  Charlotte and Emily  Bront\"e are characterized by the frequency of two words, \emph{a} and \emph{to}. In (b), the same data is visualized according to the frequency of word \emph{a} and \emph{to} in Motif type 2, i.e. ${n}_{\textrm{`a'},2}$ and ${n}_{\textrm{`to'},2}$. Finally, the partitions from Anne Bront\"e are added in (c).}\label{fig:motivation}
\end{figure}

In our analysis, each one of the considered books was split in non-overlapping partitions comprising 8,000 words each, with a total of 76 instances. For this application, we only removed punctuation marks; we did not employ any other pre-processing step.
For simplicity's sake, we illustrate the potential of the proposed technique by considering just two words in this example.
The frequency of the words \emph{a} and \emph{to} was extracted from each partition and these values are presented in Figure~\ref{fig:1.1}(a). The results show that there is a large overlapping region between Emily Bront\"e (represented by stars) and Charlotte Bront\"e (represented by circles) for the considered features. However, if one considers also the frequency of both words in motif 2 (as described in Section~\ref{characterization}), a much better discrimination can be obtained, as shown in Figure \ref{fig:1.1}(b).
This result confirms the suitability of the labelled motifs in discriminating texts, as such a good discrimination could not be obtained if only the frequency of two words were considered. The use of motifs also allows a clear distinction between Anne and Charlotte Bront\"e, though the use of these two attributes is not enough to discriminate Anne from Emily Bront\"e (see Figure \ref{fig:1.1}(c)).

In a typical authorship attribution task, the objective is to identify the author of an unknown document. For this aim, a set of documents is used to train supervised classifiers. To address this task, we firstly considered a diverse dataset comprising books written by 8 authors. This dataset, henceforth referred to as Dataset 1, is described in Table~\ref{tableDatasetI}. For the analysis, each book was truncated to the size of the shortest novel. We considered the set $W$ of the most frequent words. Therefore, the set of features consists of all combinations of words in $W$ appearing in one node of the motifs considered.
For comparison purposes, we also calculated the classification accuracies when the frequencies of the $W$ most frequent words were used as features. This frequency is calculated as the number of occurrences of each word divided by the number of tokens. Thus, it becomes possible to quantify the information gain provided by the inclusion of motifs in the traditional analysis based solely on frequency.

The classification accuracies for $|W|=\{5,10,20\}$ in Dataset 1 are presented in Table~\ref{results_aa_40}. The best results were obtained with the SVM, in general. For this reason, the discussion here is focused on the results obtained by this classifier. We note that, when comparing both versions of the proposed technique for the same $|W|$, the second version yielded best results, which reinforces the importance of function words in specific nodes. The relevance of using the local structure becomes evident when we analyze the results obtained with frequency features. While the best performance of the proposed technique reached 80\% of accuracy, the best performance obtained with frequency features was only 65\%.
\begin{table}
\caption{Accuracy rate ($\%$) in discriminating the authorship of books in Dataset 1. The best result obtained with the proposed technique surpasses by 15 percentage points the best performance obtained with traditional features based on the frequency of function words. }\label{results_aa_40}
\begin{center}
\begin{tabular}{lccccc}
 \textbf{Features}  & $|W|$ &  \textbf{J48}& \textbf{kNN}& \textbf{SVM}& \textbf{Bayes}\\
 \hline
LMV1  & 5 & 45.0	& 65.0	& 62.5	& 30.0\\
LMV1  & 10 &37.5 & 60.0	& 67.5	& 27.5\\
LMV1  & 20 & 60.0	& 65.0	& \textbf{75.0}	& 25.0\\
 \hline
 LMV2  & 5 & 55.0	& 50	.0& 62.5	& 22.5\\
 LMV2  & 10 & 47.5	& 65.0	& 77.5	& 15.0\\
 LMV2  & 20 & 45.0& 60.0	& \textbf{80.0}& 25.0\\
\hline
 MFW  & 5 &30.0	& 57.5	& 22.5	& 50.0\\
 MFW  & 10 &45.0	& 52.5	& 27.5	& 42.5\\
 MFW  & 20 &52.5	& 62.5	& \textbf{65.0}& 45.0\\
\end{tabular}
\end{center}
\end{table}

\begin{figure}
\centering
\subfigure[Labelled Motifs Version 1.\label{acc_x_mfw_1}]{\includegraphics[width=0.49\linewidth]{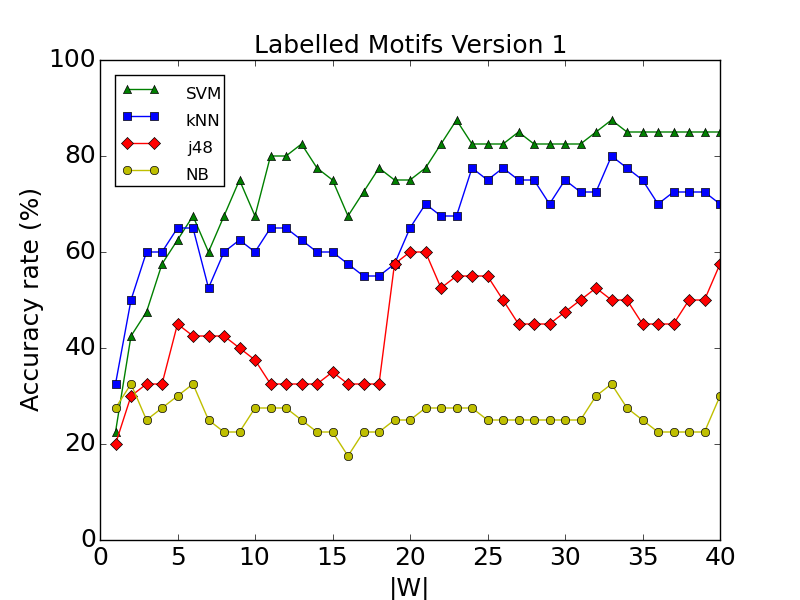}}
\subfigure[Labelled Motifs Version 2. \label{acc_x_mfw_2}]{\includegraphics[width=0.49\linewidth]{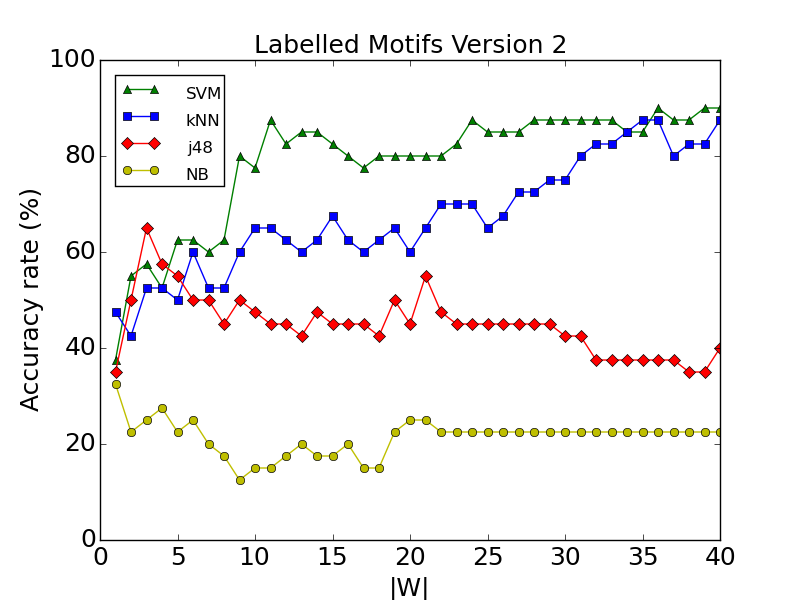}}
\caption{Accuracy rate (\%) in discriminating the authorship of books in Dataset 1 for different classifiers when several values of $|W|$ are used. While some classifiers benefit from the addition of new features, the best classifier -- the SVM -- does not require more than 20 words in $W$ to provide an excellent accuracy rate.}\label{fig:acc_x_mfw}
\end{figure}

An interesting pattern arising from the results in Table \ref{results_aa_40} is the apparent steady improvement in accuracy (for the SVM at least) as the number of features ($|W|$) increases. Therefore, we may expect that larger values of $|W|$ could yield better classification accuracies, with a corresponding  loss in temporal efficiency.
To further probe the correlation between accuracy and the value of the parameter $|W|$, we evaluated the performance of the same authorship attribution task for $1 \leq |W| \leq 40$. The percentage of books correctly classified for each value of $|W|$ is presented in Figures~\ref{acc_x_mfw_1} (LMV1) and~\ref{acc_x_mfw_2} (LMV2).
Considering the best scenario for each classifier, the SVM still outperforms all other methods. However, we did not observe an improvement in the discrimination, as the SVM does not benefit much from the addition of features. Conversely, the kNN is much benefited from the inclusion of new features. This behavior is markedly visible in the LMV2 variation, with optimized results leading to a performance similar to that obtained with the SVM. 
Based on these results and considering the efficiency loss associated with the inclusion of features, we used at most $|W| = 20$ in most of the remaining experiments.

The authorship attribution task was also evaluated using a different dataset comprising books from 9 authors, henceforth referred to as Dataset 2. This dataset is presented in Table~\ref{tableDatasetII}.
{The goal of this second experiment was to evaluate the performance of \emph{Labelled motifs} in characterizing shorter pieces of text.} In this dataset, each book was split in several non-overlapping partitions with 8,000 words each. Because some novels are longer than others, we selected the same number of partitions per author.  The classification accuracies are presented in Table~\ref{results_aa_19}. The results show, for this dataset, the best classifier for all studied features is the SVM. The best accuracies occurred for $|W|=20$. Different from the results obtained in Dataset 1, the gain in performance provided by the proposed technique is only $1.6$ percentage points. This result suggests that, in an easier authorship identification scenario, structural information plays a minor role in characterizing authors' styles. {We considered this last scenario easier for two main reasons. First and foremost, assuming that an author usually keeps his/her writing style throughout the book~\cite{Juola:2006}, we had several similar partitions extracted from each book. Therefore, the classifier was probably tested with instances very similar to others it has seen during the training phase. Second, we had fewer books per author (an average of 2.1); therefore there was less variance of writing styles per author.}

{The results obtained by the proposed technique in both authorship attribution experiments outperformed others that used only networked representations. Amancio et al.~\citep{Amancio2011a} assigned the authorship of books from a similar dataset with an accuracy rate of 65\%. When the frequency of all directed motifs with three nodes was used as attributes, Marinho et al.~\cite{Marinho2016BRACIS} achieved an accuracy of 57.5\%. In a similar study, Mehri, Darooneh and Shariati~\cite{Mehri} identified the authorship of several Persian books written by 5 authors. The authorship was correctly assigned in 77.7\% of the books. Here, our results highlight the importance of function words to characterize writing styles because most (when not all) of the words from $W$ are function words.}

\begin{table}
\caption{Accuracy rate ($\%$) in discriminating the authorship of books in Dataset 2. The best performance was obtained when the proposed technique \emph{LMV2} was used as attribute to train the SVM classifier.}\label{results_aa_19}
\begin{center}
\begin{tabular}{lccccc}
 \textbf{Features}  &  $|W|$ &  \textbf{J48}& \textbf{kNN}& \textbf{SVM}& \textbf{Bayes}\\
 \hline
LMV1  & 5 &  58.7 & 65.1 & 74.3 & 69.2\\
LMV1 	& 10 & 	61.7 & 83.7 & 91.6 & 81.3\\
LMV1	& 20 & 	66.8 & 88.3 & \textbf{95.4} & 78.7 \\
 \hline
 LMV2 	&5 &  62.1 & 67.4 & 82.0 & 69.1 \\
 LMV2 & 10&  65.5 & 80.9 & 91.6 & 75.2 \\
 LMV2 	& 20 &  68.1 & 88.0 & \textbf{96.0} & 77.5\\
\hline
 MFW  &	5&  58.3 & 67.7 & 57.8 & 73.5 \\
 MFW  & 10 &  65.8 & 83.6 & 85.3 & 83.8 \\
 MFW	& 20 &  	70.3 & 91.1 & \textbf{94.4} & 91.5 \\
\end{tabular}
\end{center}
\end{table}

\subsection{Translationese}

The term \emph{translationese} was first proposed by Gellerstam~\cite{Gellerstam}, who analyzed texts originally written in Swedish and texts translated into the same language, and concluded that the main differences between them are not related to poor translation. These differences were rather an influence of the source language on the target one.
{Several works have been dedicated to the task of translationese identification, which consists of automatically detecting original and human-translated texts~\cite{BaroniB06,vanHalteren2008,Kurokawa,IliseiIPM10,Popescu11,Koppel2011,AvnerOW16,RabinovichW16}. These methods are usually applied in a range of parallel resources, such as literary works, news articles, and transcripts from parliamentary proceedings in several languages.}

For our experiments, we selected two different datasets, the Canadian Hansard and the European Parliament. We chose them for two main reasons. First, pieces of text from the Canadian Hansard and the European Parliament debates are tagged according to the original language. Second, these translations are generally produced following good translation standards which are reflected in their quality. This makes the task of identifying the source language more challenging, providing thus another ideal scenario to probe the capabilities of the proposed methodology.

We start our analysis with data from the Canadian Hansard, which provides transcripts of debates from the Canadian parliament in the two official languages of the country, English and French. All debates are available online\footnote{http://www.parl.gc.ca/} in an XML format. During the debates, the members are allowed to speak either in English or French. Therefore, this is a rich parallel resource in which the original language of the sentences is indicated.{
Kurokawa et al.~\cite{Kurokawa} identified translationese using the 35th to 39th Parliaments from the Canadian Hansard. They analyzed the data in two granularity levels, the sentence and the block. Their blocks presented very different sizes, ranging from 3 to thousands of words. They achieved accuracies of up to 90\% with word bigram frequencies.}

In our experiment, we used 463 sessions from the 39th to 41st Parliaments, spanning the years 2006-2013. For the English side of the experiment, we divided each one of the 463 sessions into two files, one with all sentences originally produced in English (class \emph{Original}) and the other with the sentences translated into English (class \emph{Translated}). Apart from removing punctuation marks, no pre-processing step was performed in these files. We created one co-occurrence network for each file and we extracted the \emph{labelled motifs} as features for the classification. We also extracted the frequency of some of the most frequent words to compare with our results.
We proceeded in a similar way for the French side.

The results obtained with the Canadian Hansard are presented in Table~\ref{canadianResults}. The accuracies are relatively high for such a simple feature set. The results suggest that \emph{labelled motifs} are extracting information about French to English (and vice-versa) translation and these features lead to accuracies higher than the ones obtained with only the frequency of the most frequent words.
\begin{table}\caption{\label{canadianResults}Accuracy rate ($\%$) in discriminating the debates from the Canadian Hansard into two classes (\emph{Original} and \emph{Translated}). The highest accuracies were obtained with the strategy based on \emph{labelled motifs}.}
\begin{center}
\begin{tabular}{lcccccc}
 \textbf{Language} & \textbf{Features}  &  $|W|$ &  \textbf{J48}& \textbf{kNN}& \textbf{SVM}& \textbf{Bayes}\\
\textbf{English} &  & & & &\\
 \hline
& LMV1  & 20 & 90.3	& 89.2	& 96.5	& 90.3\\
& LMV2   & 20 & 90.4	& 93.5	& \textbf{97.5}	& 88.0\\
\hline
& MFW  & 5 & 71.8	& 75.0	& 57.7	& 52.7\\
& MFW  &10 & 74.4 & 	75.6	& 60.4	& 53.2\\
& MFW  &20 & 78.2 &	\textbf{79.9}	& 64.1	& 53.9\\
\textbf{French} &  & & & &\\
\hline
& LMV1 & 20 & 94.0 & 	86.5	& 97.8	& 89.4\\
& LMV2  & 20 & 94.9&	89.0	& \textbf{98.2}	& 88.9\\
\hline
& MFW & 5 & 70.2	& 69.6	& 59.0	& 53.0\\
& MFW & 10 & 71.6	& 72.1	& 56.3	& 53.3\\
& MFW & 20 & \textbf{87.1}	& 86.8	& 62.6	& 54.8\\
\end{tabular}
\end{center}
\end{table}


The ability of \emph{labelled motifs} in identify original vs. translated texts was also investigated in the
Europarl parallel corpus~\cite{europarl}, which was extracted from the Proceedings of the European Parliament. This parallel dataset includes versions in more than 20 European languages. {Similar to the Canadian Hansard, blocks of text are annotated with their original language. However, there were a few sentences with inconsistent source language tags, in which more than one language was claimed to be the source language. Those sentences were discarded in our analysis.}
For our purposes, we investigated translationese using four target languages (English, French, Italian, and Spanish) and six source languages (English, French, Spanish, Italian, Finnish, and German) from the 5th version of the corpus. Apart from removing punctuation marks, we did not employ any pre-processing step. For the English side of the experiment, we combined all sentences originally produced in English in one file (class \textit{Original}). Then, all sentences translated into English from the other five source languages (French, German, Italian, Spanish, and Finnish) were combined in one file per language (class \textit{Translated}). These 6 long files were split in non-overlapping partitions with 8,000 words each. We then selected approximately $n$ partitions from each one of the 5 source languages and $5n$ partitions from English, with $n=180$. We did this because we wanted to avoid issues with imbalanced classes. We proceeded in a similar way for the other three target languages. For French, Spanish, and Italian, we used $n=128$, $n=69$, and $n=55$ partitions, respectively.
{Here, one co-occurrence network was created for each partition. The other steps are similar to the ones applied to the Canadian Hansard dataset.}

The results obtained with the European parliament are shown in Table~\ref{europeanParliament}. To simplify our analysis, we just present the results for the classifier with the best accuracies.
Our results confirm the suitability of frequent words as relevant features, as described by Koppel and Ordan~\cite{Koppel2011}. {In a similar approach, Koppel and Ordan~\cite{Koppel2011} identified translationese in 2,000 English chunks from the Europarl corpus. They achieved an accuracy of 96.7\% using the frequency of 300 function words. However, they did not detect translationese with target languages other than English.} Once again, we have found that the characterization by \emph{labelled motifs} is extracting information about translationese. The gain in performance depends on the language being analyzed: for the English, our method surpasses the traditional one by a margin of 14 percentage points. However, for the Spanish language, the gain is only 1.4 percentage points. In the latter case, however, note that an excellent discrimination can already be obtained with the frequency of the 5 most frequent words.


\begin{table}\caption{Accuracy rate ($\%$) in discriminating the debates from the European Parliament into two classes (\emph{Original} and \emph{Translated}).}\label{europeanParliament}
\begin{center}
\begin{tabular}{lccclccc}
 \textbf{Language} & \textbf{Features} &  $|W|$ &  \textbf{SVM} & \textbf{Language} & \textbf{Features} &  $|W|$ & \textbf{SVM}\\
 \textbf{English} & & & &  \textbf{Italian} & & \\
\hline
 &LMV1  & 20 & 90.2  & & LMV1  & 20 & 93.1 \\
&LMV2  &  20 & \textbf{92.3} & & LMV2 &  20 &  \textbf{95.6}\\
\hline
&MFW  & 5 &  68.4& &MFW  &   5 & 85.2\\
&MFW  & 10 & 68.3  & &MFW  & 10 & 90.1  \\
&MFW  & 20 & \textbf{78.3} & &MFW &  20 & \textbf{92.2} \\
\textbf{French} & & & &  \textbf{Spanish} & & \\
\hline
 & LMV1  & 20 & 86.9   & & LMV1  & 20&  90.6 \\
& LMV2  & 20 & \textbf{87.6} & & LMV2  &  20 &  \textbf{92.9} \\
\hline
& MFW  & 5 & 62.0  & & MFW &   5 & 87.9\\
& MFW  & 10 &  78.0 & & MFW  & 10 & 90.4 \\
& MFW &  20 & \textbf{82.2} & & MFW  & 20 & \textbf{91.5} \\
\end{tabular}
\end{center}
\end{table}

\section{Conclusion}\label{conclusion}

{The enormous amount of texts available on the Web has increased the need for methods that automatically process and analyze this content. Therefore, several natural language processing tasks, such as authorship attribution and machine translation, have received great attention in recent years.}
In traditional approaches, texts are usually characterized by attributes derived from statistical properties of words (e.g., frequency, part-of-speech tags, and vocabulary size)~\cite{Stamatatos} and characters (e.g., frequency of characters and punctuation marks)~\cite{Grant}. In addition, syntactic and semantic features have been used as relevant attributes~\cite{Stamatatos}. More recently, interdisciplinary methodologies have also been proposed to study several aspects of texts. A well-known approach is the use of complex networks to analyze many levels of complexity of written documents. In this study, we advocated that the use of complex networks in combination with traditional features can improve the characterization of texts.

In order to combine networked methods with traditional techniques usually employed in many NLP tasks,
we proposed a hybrid method that combines the frequency of the most frequent words (mostly function words) with the occurrence of small subgraphs, called \emph{labelled motifs}. By doing so, in the context of authorship and translationese identification, we could reveal stylistic subtleties in written texts that were not extracted with only the frequency of the words. In future works, our method could be extended by considering network motifs comprising more than three nodes. Another possibility is to consider other structures particularly present in some textual networks, as paths and stars in knitted and word association networks~\cite{PalEtAl05}, respectively.


%

The results obtained in this paper suggest that the proposed approach could be applied in related tasks, such as the analysis of text complexity or the evaluation of proficiency in language learning. We believe these two tasks could be approached with our method because higher or lower complexities and proficiency levels may result in different word connections.
{Moreover, labelled motifs may also be used to detect the translation direction, i.e. given two parallel texts in different languages, which one is the original and which one was translated from the original. This information has a significant impact on statistical machine translation (SMT) systems for two main reasons. First, it has been proved that translation models trained on texts produced in the same direction of the SMT task usually perform better than the ones trained on the opposite direction~\cite{Kurokawa}. Second, translated sentences are better represented by language models compiled from translated texts~\cite{Lembersky2012}. Therefore, it is of paramount importance to automatically find the translation direction.}


\section*{Acknowledgments}
V.Q.M. and D.R.A. acknowledge financial support from S\~ao Paulo Research Foundation (FAPESP grant nos. 2014/20830-0, 2015/05676-8, 2015/23803-7 and 2016/19069-9). V.Q.M and G.H. acknowledge The Natural Sciences and Engineering Research Council of Canada.

\section*{References}

\newpage

\appendix
\setcounter{table}{0}
\renewcommand{\thetable}{A.\arabic{table}}
\section{Datasets}

\begin{table}[h]
\caption{Dataset 1 - List of 40 books written by 8 different authors.}
\label{tableDatasetI}
\begin{center}
\centering
\begin{tabular}{l|p{10cm}}
 \textbf{Author} & \textbf{Books} \\
\hline
 Arthur Conan Doyle	 & \emph{ The Adventures of Sherlock Holmes }(1892), \emph{ The Tragedy of the Korosko }(1897), \emph{ The Valley of Fear }(1914), \emph{ Through the Magic Door }(1907), \emph{ Uncle Bernac - A Memory of the Empire }(1896).\\
\hline
Bram Stoker	&\emph{ Dracula’s Guest }(1914), \emph{ Lair of the White Worm }(1911), \emph{ The Jewel Of Seven Stars }(1903), \emph{ The Man }(1905), \emph{ The Mystery of the sea }(1902).\\
\hline
Charles Dickens 	&\emph{ A Tale of Two Cities }(1859), \emph{ American Notes }(1842), \emph{ Barnaby Rudge: A Tale of the Riots of Eighty }(1841), \emph{ Great Expectations }(1861), \emph{ Hard Times }(1854).\\
\hline
Edgar Allan Poe	 &\emph{ The Works of Edgar Allan Poe, Volume 1 - 5, }(1835). \\
\hline
H. H. Munro (Saki)	 &\emph{ Beasts and Super Beasts }(1914), \emph{ The Chronicles of Clovis }(1912), \emph{ The Toys of Peace }(1919), \emph{ When William Came }(1913), \emph{ The Unbearable Bassington }(1912).\\
\hline
P. G. Wodehouse
&\emph{ Girl on the Boat }(1920), \emph{ My Man Jeeves }(1919), \emph{ Something New }(1915), \emph{ The Adventures of Sally }(1922), \emph{ The Clicking of Cuthbert }(1922).\\
\hline
Thomas Hardy 	&\emph{ A Pair of Blue Eyes }(1873), \emph{ Far from the Madding Crowd }(1874), \emph{ Jude the Obscure }(1895), \emph{ Mayor Casterbridge }(1886), \emph{ The Hand of Ethelberta }(1875). \\
\hline
William M. Thackeray  &\emph{ Barry Lyndon }(1844), \emph{ The Book of Snobs }(1848), \emph{ The History of Pendennis }(1848), \emph{ The Virginians }(1859), \emph{ Vanity Fair }(1848).\\
\end{tabular}
\end{center}
\end{table}

\begin{table}[h]
\caption{Dataset 2 - List of 19 books written by 9 different authors.}\label{tableDatasetII}
\begin{center}
\centering
\begin{tabular}{l|p{10cm}}
 \textbf{Author} & \textbf{Books}\\
\hline
 Anne Bronte &\emph{ Agnes Grey }(1847),\emph{ The Tenant of Wildfell Hall }(1848)\\
\hline

Jane Austen	&\emph{ Emma }(1815),\emph{ Mansfield Park }(1814),\emph{ Sense and Sensibility }(1811)  \\
\hline

Charlotte Bronte &\emph{ Jane Eyre }(1847),\emph{ The Professor }(1857)\\
\hline\setcounter{table}{0}
\renewcommand{\thetable}{A\arabic{table}}

James Fenimore Cooper	&\emph{ The Last of the Mohicans }(1826),\emph{ The Spy }(1821),\emph{ The Water Witch }(1831) \\
\hline

Charles Dickens	&\emph{ Bleak House }(1853),\emph{ Dombey and Son }(1848),\emph{ Great Expectations }(1861) \\
\hline

Ralph Waldo Emerson &\emph{ The Conduct of Life }(1860),\emph{ English Traits }(1853) \\
\hline

Emily Bronte &\emph{ Wuthering Heights }(1847)\\
\hline

Nathaniel Hawthorne	&\emph{ The House of the Seven Gables }(1851)\\
\hline

Herman Melville &\emph{ Moby Dick }(1851),\emph{ Redburn }(1849)\\

\end{tabular}
\end{center}
\end{table}







\end{document}